\documentclass[journal]{IEEEtai}

\usepackage[colorlinks,urlcolor=blue,linkcolor=blue,citecolor=blue]{hyperref}

\usepackage{color,array}

\usepackage{graphicx}

\usepackage{cite}
\usepackage{amsmath,amssymb,amsfonts}
\usepackage{algorithmic}
\usepackage{graphicx}
\usepackage{textcomp}
\usepackage{xcolor}
\usepackage{comment}
\def\BibTeX{{\rm B\kern-.05em{\sc i\kern-.025em b}\kern-.08em
		T\kern-.1667em\lower.7ex\hbox{E}\kern-.125emX}}

\begin{document}

\title{Hyperspectral Imaging-Based Grain Quality Assessment With Limited Labelled Data}

\author{Priyabrata Karmakar, Manzur Murshed, \IEEEmembership{Senior Member, IEEE} and Shyh Wei Teng
%\thanks{This paragraph of the first footnote will contain the date on which you submitted your paper for review. It will also contain support information, including sponsor and financial support acknowledgment. For example, ``This work was supported in part by the U.S. Department of Commerce under Grant BS123456.'' }
\thanks{ Priyabrata Karmakar is with the School of Information Technology, Deakin University, Australia (e-mail: priyabrata.karmakar@deakin.edu.au).}
\thanks{Manzur Murshed is with the School of Information Technology, Deakin University, Australia (e-mail: m.murshed@deakin.edu.au).}
\thanks{Shyh Wei Teng is with the Institute of Innovation, Science and Sustainability, Federation University, Australia (e-mail: shyh.wei.teng@federation.edu.au).}
%\thanks{This paragraph will include the Associate Editor who handled your paper.}
}

\markboth{Journal of IEEE Transactions on ABC, Vol. 00, No. 0, Month 2020}
{Priyabrata Karmakar \MakeLowercase{\textit{et al.}}: Bare Demo of IEEEtai.cls for IEEE Journals of IEEE Transactions on ABC}

\maketitle

\begin{abstract}
Recently hyperspectral imaging (HSI)-based grain quality assessment has gained research attention. However, unlike other imaging modalities, HSI data lacks sufficient labelled samples required to effectively train deep convolutional neural network (DCNN)-based classifiers. In this paper, we present a novel approach to grain quality assessment using HSI combined with few-shot learning (FSL) techniques. Traditional methods for grain quality evaluation, while reliable, are invasive, time-consuming, and costly. HSI offers a non-invasive, real-time alternative by capturing both spatial and spectral information. However, a significant challenge in applying DCNNs for HSI-based grain classification is the need for large labelled databases, which are often difficult to obtain.
To address this, we explore the use of FSL, which enables models to perform well with limited labelled data, making it a practical solution for real-world applications where rapid deployment is required. We also explored the application of FSL for the classification of hyperspectral images of bulk grains to enable rapid quality assessment at various receival points in the grain supply chain. We evaluated the performance of few-shot classifiers in two scenarios: first, classification of grain types seen during training, and second, generalisation to unseen grain types, a crucial feature for real-world applications. In the first scenario, we introduce a novel approach using pre-computed collective class prototypes (CCPs) to enhance inference efficiency and robustness. In the second scenario, we assess the model's ability to classify novel grain types using limited support examples. Our experimental results show that despite using very limited labelled data for training, our FSL classifiers accuracy is comparable to that of a fully trained classifier trained using significantly larger labelled database. We also propose a novel enhancement to the squeeze and excitation attention mechanism to improve feature representation in hyperspectral images. These findings demonstrate the potential of FSL as a practical solution for rapid, accurate grain quality assessment in real-world applications.
\end{abstract}

\begin{IEEEImpStatement}
This study combines hyperspectral imaging (HSI) with few-shot learning (FSL) to address the need for large labelled datasets in grain quality assessment. By introducing collective class prototypes (CCPs) and enhancing feature representation with a novel attention mechanism, the approach enables accurate classification with minimal labelled data. It performs well on both seen and unseen grain types, making it practical for real-world applications like rapid grain quality checks in supply chains. This method offers a fast, non-invasive, and efficient alternative to traditional assessments, benefiting the grain industry significantly.
\end{IEEEImpStatement}


 \section{Introduction}

\IEEEPARstart{D}{ifferent} types of grains, like wheat, rice, and corn serve as staple foods worldwide. Therefore, effective and efficient assessment of grain quality is significantly important, especially in grain trading process and in food safety assurance. Traditionally, grain quality assessments involve chemical and biological analysis. However, these methods are invasive, destructive, time-consuming, and costly. Consequently, there is a shift in the testing methodology towards faster, non-invasive, non-destructive, and real-time approaches.

Hyperspectral imaging (HSI) is a promising non-invasive and real-time method for assessing grain quality. HSI has been applied successfully in agriculture and the food industry, such as in detecting damage in fruits and vegetables \cite{ariana2006near, zhu2019rapid}, contaminants in food \cite{he2015hyperspectral}, and quality in dairy \cite{calvini2020exploring} and meat \cite{feng2020colour}. HSI is increasingly used for evaluating grain quality parameters like protein \cite{shuqin2016predicting}, moisture \cite{mahesh2011identification}, defects \cite{barbedo2018detection}, and contaminants \cite{liang2020comparison}.

Grain classification, an important part of quality assessment, has traditionally relied on invasive methods \cite{kniese2021classification}. From the early 2000s, non-invasive techniques like X-ray imaging \cite{neethirajan2007detection}, near-infrared spectroscopy (NIRS) \cite{bao2019rapid}, and RGB imaging with shape, color, and texture features \cite{majumdar2000classification, pourreza2012identification} became more common. However, X-ray imaging poses health risks, and NIRS and RGB imaging have limitations in leveraging both spatial and spectral information.

HSI addresses these limitations by combining NIRS and digital optical imaging to capture both spatial and spectral data as three-dimensional (3D) hypercubes with two spatial and one spectral dimension \cite{gowen2007hyperspectral}. This provides comprehensive data, revealing patterns based on unique interactions of electromagnetic (EM) energy with biological materials, which vary by chemical composition and structure.

In recent years, deep convolutional neural networks (DCNNs) have gained popularity in image processing, excelling at recognising complex patterns in raw data. Originally used for RGB imaging, DCNNs are now widely applied to multispectral and HSI data. Traditionally, DCNNs in HSI have focused on single kernels or sparse samples \cite{pub.1064892569, qiu2018variety, weng2020hyperspectral}, but this approach is inefficient and time-consuming \cite{karmakar2023guide}. Bulk HSI imaging of densely distributed kernels, however, expedites the process \cite{ravikanth2016detection, shao2020determination}, making it ideal for fast quality assessments in the grain supply chain, where time is crucial. Bulk HSI imaging enhances efficiency, surpassing traditional methods in speed and streamlining grain quality evaluation.

The authors of \cite{dreier2022hyperspectral} proposed a DCNN-based approach for analysing and classifying bulk grains using 3D hyperspectral data cubes. They have successfully performed grain classification with hyperspectral datacubes using a two-dimensional (2D) DCNN (specifically, ResNet) that is traditionally used to work with RGB images after incorporating a small modification of adding a linear down-sampling layer at the beginning of ResNet to utlise the spectral data. 

One of the main challenges of using HSI for any application is the availability of a large enough labelled database. This is because, acquiring accurate and comprehensive ground truth data for hyperspectral targets is labour-intensive due to the complexity of the data. This requires precise annotation and validation, often involving expert knowledge and manual effort. Furthermore, the sheer volume of hyperspectral information complicates the creation of an extensive ground truth database \cite{zhao2019laboratory,nalepa2021recent}. In addition, in the specific application scenarios the scarcity of enough labelled data can be witnessed. For instance, in a grain supply chain, where often new grades of an existing grain are introduced and their qualities needs to assessed with the similar accuracy as the existing grades. 

Therefore, researchers often must work with only a small ground truth database. However, achieving satisfactory results using DCNN-based methods, such as in \cite{dreier2022hyperspectral}, requires a significantly larger labelled database, comparable in terms of the number of images to the RGB database typically used for similar DCNN-based approaches.

\begin{comment}
	The issue of working with smaller labelled databases can be overcome by using FSL approaches. In the RGB imaging domain, the success of FSL is evident \cite{ren2024few}. FSL has been explored in the HSI domain but mostly restricted to hyperspectral images where a single hysperspectral image contains multiple classes (i.e., labels) and each pixel in the hyperspectral image belongs to an individual label \cite{bai2022few}. However, the application of using FSL on hyperspectral images of bulk grains where each image belongs to a single class is limited. Therefore, in this paper, we aim to investigate the performance of FSL for classifying hyperspectral images of bulk grains. Specifically, we aim to investigate whether and how effective different configurations of FSL in classifying hyperspectral images of bulk grains.
	
	In the context of few-shot image classification, a model is typically constructed using a pre-trained DCNN as a backbone. This approach leverages transfer learning principles, allowing the model to benefit from features learned on large-scale databases while adapting to the specific task at hand with limited labelled examples.
	One of the primary motivations for employing FSL in HSI is to develop models capable of learning robust feature representations within this specialised domain. By training on a diverse yet limited set of hyperspectral images, the model can potentially capture the underlying spectral and spatial characteristics unique to this imaging modality.
\end{comment}

Few-shot learning (FSL) can help address the challenge of small labelled databases. It has shown success in the RGB imaging domain \cite{ren2024few}. FSL has also been applied to hyperspectral imaging (HSI), though mainly for cases where a single hyperspectral image includes multiple classes, with each pixel labelled individually \cite{bai2022few}. However, its use in classifying bulk grain hyperspectral images, where each image represents a single class, is limited. This paper investigates FSL's performance in classifying bulk grain hyperspectral images and explores different FSL configurations for this purpose.

In few-shot classification, a pre-trained DCNN backbone is often used \cite{dvornik2019diversity}. This allows the model to leverage features from large databases and adapt to specific tasks with minimal labelled examples. FSL in HSI aims to create models that learn robust feature representations by training on a diverse but small set of hyperspectral images, capturing unique spectral and spatial characteristics.

FSL models, particularly in the context of hyperspectral image classification, offer significant advantages beyond their ability to effectively classify unseen (test) images from the classes they were trained on. A key strength of these models lies in their capacity to generalise to novel classes without requiring complete retraining of the classifier. This capability is especially valuable in domains like HSI, where data acquisition and labeling can be resource-intensive.

Once a few-shot model has been trained on a specific hyperspectral domain, it can adapt to classify new classes of hyperspectral images captured under similar conditions by utilising a small support set of examples for these novel classes \cite{li2023deep}. This adaptability is achieved through meta-learning techniques, where the model learns to learn from few examples, rather than learning specific class features \cite{lee2024unlocking}.

In grain classification, for instance, an FSL model trained on wheat images can quickly adapt to identify other grains like rye, rice, or corn. This is especially useful when few labelled images are available. Unlike traditional methods, which are slow and data-intensive, an FSL model enables fast deployment for classifying different grades within new grain types using minimal labelled data \cite{finn2017model}.

This approach is also beneficial when a new grade of grain needs to be identified with an existing classifier. A standard classifier would need re-training to recognise the new grades. However, a classifier trained with 
FSL can quickly adapt using only a small set of labelled samples. New grain grades often appear due to genetic advancements, market demand, or seasonal quality variations \cite{new_grade_arrival, abdul2022genetically}. In supply chains, quality checks on new grades must match the standards for existing ones. But gathering labelled images and re-training is time-consuming. It reduces assessment efficiency, defeating the purpose of using digital methods instead of traditional, time-consuming approaches. An FSL-trained classifier avoids this by adapting without extensive re-training or large labelled databases.

The FSL architecture comprises support and query sets, with support sets serving as references for classifying images in query sets \cite{luo2023closer}. One significant challenge of using FSL with hyperspectral images is the existence of outliers \cite{pal2022few}. In the context of HSI, the likelihood of encountering outliers in support sets is significantly higher compared to RGB imaging. This is primarily due to the high dimensionality of hyperspectral data cubes and the complex, often unknown relationships among different spectral channels \cite{zheng2023fusion}. When creating support sets from such high-dimensional databases, the increased variance among samples naturally leads to a higher probability of outlier occurrence \cite{wang2024hypersigma}. While this may not be a significant issue in RGB imaging, it presents a considerable challenge in the HSI domain. The problem gets even worse when assessing grain quality using hyperspectral images captured in various ad hoc environments, such as different receival points along a supply chain, rather than in controlled laboratory settings. These non-ideal conditions can introduce noise and artifacts into the captured images \cite{xu2021motion}, resulting in increased intra-class variance within support sets and a higher probability of outlier presence. This phenomenon poses significant challenges for accurate classification and analysis of hyperspectral data, particularly in real-world applications. Therefore, the high probability of outlier existence needs to be addressed algorithmically to avoid performance loss. We have aimed to address this issue in this paper by proposing collective class prototypes (CCP). This proposal is discussed in detail in Section IV.

%The FSL architecture includes support and query sets, where support sets act as references for classifying query images \cite{luo2023closer}. A major challenge in applying FSL to hyperspectral images (HSI) is the high chance of outliers in support sets \cite{pal2022few}, mainly due to the high dimensionality and complex spectral relationships in hyperspectral data \cite{zheng2023fusion}. This variance increases the likelihood of outliers, a problem less common in RGB imaging but critical in HSI, especially when assessing grain quality using hyperspectral images captured in various ad hoc environments, such as different receival points along a supply chain, rather than in controlled laboratory settings. These conditions can introduce noise and artifacts, raising intra-class variance and outlier risk \cite{xu2021motion}, which complicates accurate classification. To address this, we propose collective class prototypes (CCP), detailed in Section IV.

In this paper, we investigate the performance of a trained few-shot classifier in two scenarios: first, classifying test images from the same classes on which the classifier was trained, and second, classifying images from novel and unseen classes that were not present during the training phase. We also aim to investigate the performance of the few-shot classifier compared to a fully trained classifier. Since few-shot classifiers use much fewer labelled images for training, our goal is to get the few-shot classifier's accuracy as close as possible to that of a fully trained classifier.%  but with relatively significantly greater efficiency.
Specifically, in this paper, our contributions are as follows.

\begin{itemize}
	\item We explored two FSL scenarios for hyperspectral image classification to assess grain quality. %at various supply chain receival points. 
	Scenario 1 simulates supervised classification, where the classifier is trained on classes it will predict during inference. Scenario 2 evaluates the classifier's ability to predict unseen classes, given a support set.
	\item In the first scenario, we propose using pre-computed CCPs in place of support sets during inference. By computing the average of prototypes representing each class, CCPs provide a more robust representation of classes and are less susceptible to outliers that may be present in individual support sets. This is because, averaging helps to align reference data with a central tendency while minimising the impact of outliers. This approach also reduces inference time, as it eliminates the need to extract features from the support set at each runtime.
	\item In the second scenario, we aim to evaluate the classifier's ability to generalise to unseen classes, which is crucial in supply chain applications where new grain grades or types may be encountered. We seek to quantify the classifier's performance in classifying images from unseen classes, where limited time and data availability make it impractical to train a new classifier.
	\item  We investigated the effectiveness of our few-shot image classifiers, which require significantly less training data than fully trained or fine-tuned supervised classifiers. We evaluated how closely our few-shot classifier's accuracy is to that of a fully trained supervised classifier.
	\item We explored how attention mechanism can improve the hyperspectral feature representation. Specifically, we proposed a novel squeeze mechanism within the squeeze and excitation attention mechanism to enhance feature representation of hyperspectral images.

\end{itemize}

The rest of paper is organised as follows. Sections II and III provide a brief introduction to FSL and a brief introduction to squeeze and excitation attention mechanism, respectively. Section IV presents the proposed methodology. Section V outlines the experimental details. Section VI discusses the results. Finally, Section VII concludes the paper.

%The bulk imaging of grain kernels expedites the assessment process and it is beneficial where time is limited. For example, at different receival points in a grain supply chain, HSI on bulk samples will make the quality assessment process faster compared to HSI on sparsely distributed kernels let alone compared to using traditional invasive approaches.

\section{Brief Introduction to Few-Shot Learning}
%FSL (FSL) is a machine learning framework that enables a pre-trained model to generalize over new categories of data with only a few examples. It is particularly useful in scenarios where acquiring large amounts of labelled data is impractical or costly. FSL has gained significant attention in the research community due to its potential to enhance the flexibility and generalization capabilities of machine learning models across various applications. 

%In the context of FSL, there are several sub-fields, including N-shot learning, which involves discriminating between N classes using a few examples, and meta-learning, where the learning algorithm improves itself through multiple training episodes. Approaches to FSL problems are discussed from the perspectives of meta-learning, transfer learning, and hybrid approaches.

%One of the key approaches to FSL is the prototypical network. The prototypical network is a few-shot classification model that learns a metric space where classification can be performed by computing distances to prototype representations of each class. This approach has shown promising results in enabling machine learning models to learn from a small number of examples, thus reducing the dependency on large labelled databases and enhancing the adaptability of models to new tasks and domains.

FSL is a sub-field of machine learning that addresses the challenge of training models with a small number of labelled examples, making it useful in scenarios where acquiring large amounts of labelled data is impractical or costly. It aims to enable pre-trained machine learning models to generalise over new categories of data, based on a limited number of samples, thus reducing the dependency on large labelled databases. It belongs to the realm of meta-learning, which involves the concept of learning to learn. \cite{wang2020generalizing}.

The concept of FSL is discussed as follows. FSL consists of support and query sets and an $N$-way $K$-shot learning scheme. 

\begin{itemize}
	\item{Support set:} The support set contains a limited number of labelled samples (i.e., samples and their corresponding ground truth labels) for each class of data under consideration. These samples are utilised by a pre-trained model to generalise and adapt to the new classes.
	
	\item{Query set:} The query set contains unlabelled samples from the same classes. The model applies the knowledge gained from the support set to make predictions on the query set to assess its ability to generalise to unseen new classes.
	
	\item{$N$-Way $K$-shot learning scheme: } $N$-way denotes the presence of $N$ novel classes that a pre-trained model must generalise to. $K$-shot specifies the number of labelled samples that exist in the support set for each of the $N$ new classes. If the value of K is very small, it means the number of supporting samples in each class is much less and it causes the few shot task to be very difficult.
\end{itemize}

In the literature, there are three types of FSL can be found and they are metric-based, optimisation-based, and model-based \cite{parnami2022learning}. Among them, metric-based FSL offers more advantages due to its capacity to explicitly learn distance metrics, providing flexibility across varied tasks and demonstrating superior performance and generalisation compared to optimisation-based and model-based approaches.

The metric-Based FSL focuses on learning distance metrics for effective generalisation with limited data. Some of the popular metric-based FSL approaches include Siamese networks, Matching networks, and Prototypical networks \cite{kulismetric}. In this paper, we focus on Prototypical networks because they demonstrate superior performance by learning class prototypes, allowing for robust representations and accurate few-shot classification in diverse scenarios \cite{snell2017prototypical}. 

\subsection{Prototypical Network}

%The Prototypical networks work based on the extracted prototypes. These prototypes are centroids in an embedding space, where data points are mapped to capture essential class characteristics. A distance function is used to calculate the distances between the prototypes and compare them based on the calculated distances.

%Prototypical networks operate by learning a feature space using an embedding architecture. This embedding architecture extracts features from the raw data (i.e., image in the context of this paper) to map it to the corresponding feature space. In this space, classification is executed by computing distances to prototype representations of each class. The networks achieve this by computing a distinctive representation, or prototype, for each class through an embedding function. This embedding function calculates the mean vector of the embedded support points that belong to a particular class. When a new image is to be classified, its embedding is computed, and its distance to each class prototype is measured. The image is then assigned to the class whose prototype is closest in the learned feature space. A mathematical explanation of Prototypical networks is provided in the following paragraphs.

Prototypical networks classify images by learning a feature space through an embedding architecture which extracts features from raw data (like images) and maps them to this space. Each class has a prototype, represented by the mean vector of its support points' embeddings. To classify a new image, the network computes its embedding and measures the distance to each class prototype. The image is assigned to the class with the closest prototype. A detailed mathematical explanation of Prototypical networks is provided in the following paragraphs.

Say, support set $set_{Support} = \{ (x_1, y_1), \dots, (x_N, y_N) \}$ consists of $N$ number of  images ($x$) and their corresponding labels ($y$) and there are total $K$ different classes. The value of $N$ is significantly lower compared to a typical labelled image database used in standard image classification applications. The operation of Prototypical networks starts with computing prototypes of each class through an embedding function ($f_{\phi})$ that is often a pre-trained CNN. Each prototype is derived from the mean vector of the embedded support points associated with its respective class given by \eqref{eq:prototype}

\begin{equation}
	c_k = \frac{1}{\lvert S_k \rvert} \sum_{(x_i, y_i) \in S_k} f_\phi(x_i),
	\label{eq:prototype}
\end{equation}
where $S_k$ and  $c_k$ are the set of samples and the computed prototype, respectively of class $k$. 

Now, using a distance function $d$, Prototypical networks generate a distribution over classes for a test image $x_q$ (image from the query set)  by employing a softmax function over the distances to the prototypes in the embedding space. Equation \eqref{eq:probability} shows the probability of a test image belonging to a class $k$.

\begin{equation}
	p_\phi(y = k | x) = \frac{e^{-d(f_\phi(x), c_k)}}{\sum_{k'} e^{-d(f_\phi(x), c_{k'})}}, 
	\label{eq:probability}
\end{equation}
where $k' \in K$ represent each classes in $K$.

During the training phase, the network learns by minimising the negative log-probability ($ -p_\phi(y = k | x) $) of the true class $k$ via stochastic gradient descent (SGD). From the training set, a subset of classes is selected randomly, and out of that again a subset is selected as a support set, and the rest is considered to use as a query set.

\section{Brief Introduction to Squeeze and Excitation attention}
In this section, we discuss the Squeeze and Excitation (SE) attention mechanism \cite{hu2018squeeze}. It has three components. They are as follows.

\begin{itemize}
	\item Squeeze: In this step, the spatial dimensions of each channel are reduced to a single value which is the global representation of the corresponding channel.  This is done using a pooling operation, like global average pooling (GAP). Mathematically, the squeezed representation, $z_c$ of each channel $c$ is given by \eqref{eq:squeeze}.
	
	\begin{equation}
		z_c = \frac{1}{H \times W} \sum_{i=1}^{H} \sum_{j=1}^{W} I(i, j, c),
		\label{eq:squeeze}
	\end{equation}
	where $H$, $W$, and $C$ represent the height, width and the total number of spectral channels of each hyperspectral image, $I$.
	
	\item Excitation: In the excitation step, the SE block uses two fully connected (FC) layers to capture channel-wise dependencies and relationships. This step computes attention weights for each channel based on the squeezed global values. At first, the data from the last step is passed through a reduction layer that reduces the number of channels (by a factor called the reduction ratio). This introduces non-linearity and helps the model learn complex relationships between channels. After that, the reduced representation is expanded back to the original number of channels, producing a set of attention weights through a sigmoid activation function. These attention weights are in the range (0, 1), indicating the importance of each channel. The mathematical model of this step is given by \eqref{eq:excitation}.
	
	\begin{equation}
		s = \sigma \left( W_2 \cdot \text{ReLU} \left( W_1 z \right) \right),
		\label{eq:excitation}
	\end{equation}
	where $W_1$, $W_2$ and $\sigma$ are the weights of the first (reduction) and second (expanding) fully connected layers and $sigmoid$ function, respectively.
	
	\item Recalibration: This is the final step of SE block. In this step, given by \eqref{eq:recalibration} learned attention weights are used to rescale the original channel data by multiplying each channel by its corresponding attention weight. This allows the network to emphasise the important channels and suppress the less relevant ones.
	
	\begin{equation}
		I^{'}_c = {I}_c \cdot s_c,
		\label{eq:recalibration}
	\end{equation}
	where $s_c$, $I_c$ and $I^{'}_c$ are the attention weights, original  data, and scaled data of the channel $c$, respectively.
\end{itemize}

\section{Proposed Methodology}

%In the proposed modeling approach, we have considered the Prototypical network as our foundation model. In addition,  restnet18 and resnet50 are separately considered as the backbone model to the Prototypical network. For FSL, 5-shot and 10-shot configurations are considered. The performance of the Prototypical network is recorded both before and after the network training. After training, the inference is performed in two approaches: a. fewshot approach (similar to before training but with a trained network) b. zero-shot approach. 

%In our proposed  methodology, we have considered the Prototypical network as the fundamental architecture, employing the Euclidean distance function ($L2-norm$) to compute the pairwise distances between embeddings. A state-of-the-art 2D CNN is considered as backbone for embedding the hyperspectral images to map them in the feature space.  In \cite{dreier2022hyperspectral}, for grain classification using HSI, authors preferred a 2D CNN over a 3D one. They proposed modifying a 2D ResNet to handle spectral data by adding a spectral down-sampling layer at the beginning. 

In our proposed  methodology a Prototypical network with Euclidean distance ($L2$-norm) to calculate pairwise distances between embeddings is considered. A state-of-the-art 2D CNN serves as the backbone for embedding hyperspectral images into feature space. For grain classification with HSI, \cite{dreier2022hyperspectral} favours a 2D CNN over a 3D one, modifying a 2D ResNet to handle spectral data by adding a spectral down-sampling layer. This allowed the network to quickly reduce the spectral dimension while still utilising spectral features. This modification increased the efficiency in model training compared to using a 3D CNN. In addition, backed by the experimental results, the authors of \cite{dreier2022hyperspectral} also claimed that this modification achieved better accuracy. A potential explanation for this enhancement is that the rapid reduction of spectral dimensions helped evade the overfitting problem, allowing the model to generalise better. Therefore, it was found to be more efficient in leveraging both spectral and spatial information. 

\begin{figure}
	\centering
	\includegraphics[width=1\linewidth]{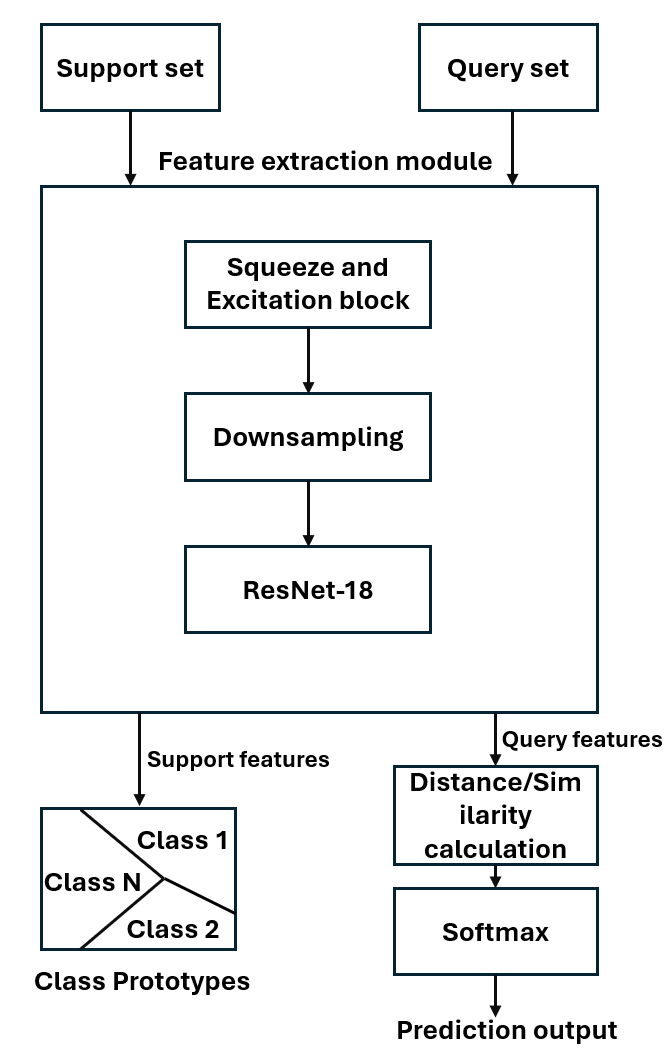}
	\caption{Block diagram of Proposed Methodology}
	\label{fig:block-diagram}
\end{figure}

This motivated us to use a 2D ResNet (following the same modification of adding an spectral down sampling layer at the beginning) as the backbone for our experiments. In addition, we incorporated a channel-wise attention mechanism to emphasise informative spectral bands and reduce noise before spectral downsampling. The mechanism utilised a squeeze-and-excitation block placed before the downsampling layer. A block diagram of our proposed approach is shown in Fig. \ref{fig:block-diagram}.

%A SE block is a neural network module designed to improve the representational power of CNNs by adaptively recalibrating the channel-wise feature information \cite{hu2018squeeze}. In general, the SE block mainly aims to enhance the network's ability to focus on the most important feature channels encoded by a deep CNN in the RGB imaging space \cite{naresh2024empirical}. In the existing literature, the SE block has been explored in the HSI domain, but its application has been primarily constrained to re-calibrating features derived from CNN embeddings, rather than directly operating on the raw HSI data cubes \cite{nguyen2024hyperspectral, wang2023advances}. However, this approach may not fully leverage the rich spectral information contained in the raw HSI data, this limits its effectiveness. Therefore, in our proposal, we used SE block to assign higher weight to the important spectral bands while suppressing the less informative ones. It does this by learning attention weights for each channel. This helps the model emphasise the more relevant channels. 

An SE block enhances CNNs by recalibrating channel-wise features to focus on the most important ones \cite{hu2018squeeze}. It improves the network's ability to highlight key feature channels, particularly in the RGB imaging space \cite{naresh2024empirical}. While SE blocks have been applied in the HSI domain, their use has mostly been limited to recalibrating CNN-derived features rather than directly working on raw HSI data cubes \cite{nguyen2024hyperspectral, wang2023advances}. This limits their ability to fully utilise the rich spectral information in raw HSI data. To address this, our approach uses the SE block to assign greater weight to significant spectral bands while suppressing less informative ones. By learning attention weights for each channel, the model focuses on the most relevant channels.

%In this paper, we have modified the squeeze step of the SE block by combining adaptive average pooling and adaptive max pooling given by \eqref{eq:proposed_squeeze}, rather than using only average pooling. The average pooling captures global trends by smoothing feature activations but can fail to highlight key spectral features \cite{zhong2017spectral}, while max pooling retains prominent activations but may miss the broader context \cite{zhao2024improved}. By averaging the outputs of both pooling methods, we aimed to balance the global feature aggregation with identifying strong activations. This has produced more discriminative attention weights across spectral channels. This approach enhances feature representation and it leads to improved classification performance in HSI.

In this paper, we modified the squeeze step of the SE block by combining adaptive average pooling and adaptive max pooling as described in \eqref{eq:proposed_squeeze}, instead of using only average pooling. Average pooling smooths feature activations to capture global trends but may overlook key spectral features \cite{zhong2017spectral}, while max pooling highlights strong activations but can miss broader context \cite{zhao2024improved}. Combining both methods balances global feature aggregation and prominent activation detection, resulting in more discriminative attention weights across spectral channels. This improved feature representation enhances HSI classification performance.

\begin{equation}
	z_c = \frac{1}{2} \left( \frac{1}{H \times W} \sum_{i=1}^{H} \sum_{j=1}^{W} I(i, j, c) + \max_{i,j} I(i, j, c) \right).
	\label{eq:proposed_squeeze}
\end{equation}

Obtaining sufficient samples for hyperspectral image classification is challenging, making it difficult to build a fully trained or fine-tuned classifier. To address this, we adopted a few-shot classification approach. During evaluation, the few-shot classifier requires a labelled support set to compare and determine the grade of a grain sample. However, providing a support set during evaluation has two key limitations. They are as follows.

\begin{comment}

	\begin{itemize}
		\item The classifier (here, the Prototypical network) needs to extract the features of images from support set along with the test image (i.e, query), every time it requires to assess grades of the grain samples. Extracting features of the support set images is time-consuming and it increases the time complexity in proportion to the shot-size. Lower shot-size will reduce the time complexity but it may compromise with the effectiveness as lower shot-size support set may not effectively represent the different classes.
		\item The support set images may contain some images that are outliers and it may make the mean features (i.e., prototypes) of the support set biased. This can reduce the classification accuracy.
	\end{itemize}
	
\end{comment}

\begin{itemize}
	% \item The classifier, in this case the Prototypical network, must extract features from both the support set images and the test image (query) each time it is required to assess the grades of grain samples. Extracting features from the support set can be time-consuming, and the time complexity increases with the shot size. While reducing the shot size can lower time complexity, it may come at the cost of classification effectiveness, as a smaller support set may not adequately represent the different classes.
	
	\item The classifier, in this case the Prototypical network must extract features from both the support set and the query image each time it evaluates grain grades. This process can be time-consuming, especially as the shot size increases. Reducing the shot size lowers time complexity but may weaken classification accuracy, as a smaller support set might not fully represent the classes.
	
	\item The support set may contain outlier images, which could bias the mean feature representations (prototypes) of the classes. This bias can negatively impact classification accuracy.
\end{itemize}

%To overcome the above limitations, in this paper we propose collective class prototypes (CCP). After the Prototypical network is trained, the prototypes of each class of each support set in the training set are computed. These prototypes are then averaged to compute a collective prototype for each class. These are the CCPs given by \eqref{eq:avg_prototype}. They belong to the same feature space as the features extracted using the trained Prototypical network with the fine-tuned embedding model. The CCPs are then utilised for classification by measuring the distance with the feature representation of test images. 

To address these limitations, we propose CCP. After training the Prototypical Network, we calculate the prototypes for each class within the training set's support sets. These prototypes are averaged to create a collective prototype for each class, as defined in \eqref{eq:avg_prototype}. The CCPs, residing in the same feature space as the extracted features from the fine-tuned embedding model, are then used for classification by measuring their distance to the test images' feature representations.

\begin{equation}
	C_k = \frac{1}{N} \sum_{i\in N} c_k^i,
	\label{eq:avg_prototype}
\end{equation}
where $C_k$ is the CCP and $c_k$ is the prototype at the $i^{th}$ episode of class $k$. $N$ and $i$ represent the total episodes and individual episode, respectively in the best training iteration.

%These CCPs accompany the trained Prototypical network wherever the test images are required to be evaluated. Whether it is at the site of supply chain receival points or hosting them at the cloud to access via an Application Programming Interface (API).  By eliminating the need for support sets during evaluation, CCPs not only significantly improve efficiency but also enhance effectiveness. This is because the collective nature of CCPs, which are derived by averaging multiple prototypes, reduces the likelihood of bias toward any individual outlier. The averaging process mitigates the influence of any atypical or noisy examples within the training data, thereby yielding more robust and generalised class representations for classification.

These CCPs accompany the trained Prototypical network for evaluating test images, whether at supply chain receival points or hosted on the cloud for application programming interface (API) access. By removing the need for support sets during evaluation, CCPs improve both efficiency and effectiveness. Their collective nature, derived from averaging multiple prototypes, reduces bias from outliers. This averaging process minimises the impact of noisy or atypical training examples, resulting in more robust and generalised class representations for classification.

\section{Experiment}

\subsection{Database}
In this paper, we have considered the HSI database used in \cite{dreier2022hyperspectral}. This database consists of hyperspectral images of eight grains harvested in Denmark and Sweden. They are as follows. Midsummer Rye (Rye), Spelt wheat (Spelt), Halland wheat (Halland), \O land wheat (Oland), Winter wheat, Type A - Sweden (WH 1), Spring wheat (WH 3), Winter wheat, Type A - Denmark (WH 4), Winter wheat, Type B (WH 5). A brief description of these grains provided in Table \ref{tab:grains}.

\begin{table}
	\caption{Brief description of the grains}
	\centering
	\begin{tabular}{p{0.7cm}|p{1.8cm}|p{1cm}|p{1.5cm}|p{1.5cm}}
		\hline
		Label & Grain type  & Country & Average protein content (\%) & Average moisture content (\%) \\
		\hline
		Rye & Midsummer Rye & Denmark & 10.42 - 10.73  & 10.50 - 10.60\\
		
		Spelt & Spelt wheat  & Denmark  & 13.00 - 13.37  & 10.65 - 10.95\\
		
		Halland & Halland wheat & Denmark & 14.70 - 14.90 & 10.70 - 10.80\\
		
		Oland & \O land wheat & Denmark & 14.70 & 10.85 - 11.95\\
		
		WH 1 & Winter wheat, Type A & Sweden & 12.10 - 12.23 & 11.60 - 12.55\\
		
		WH 3 &  Spring wheat &  Sweden & 12.90 - 13.06 & 11.70 - 12.30\\
		
		WH 4 & Winter wheat, Type A  & Denmark & 11.60 - 11.74 & 12.15 - 13.65\\
		
		WH 5 & Winter wheat, Type B & Sweden & 10.50 - 10.62  & 11.80 - 12.35 \\
		\hline
	\end{tabular}
	
	\label{tab:grains}
\end{table}

%The images were captured in dense and sparsely packed settings. The imaging camera  (Specim FX17 line scan camera \cite{specim}) features 640 spatial pixels and operates within the 900–1700 nm range, detecting 224 spectral channels evenly distributed across its spectrum. For all images, the initial 10 and final 10 spectral channels of the hyperspectral image data were excluded due to low camera sensitivity and increased noise in this range. As a result, the data used for further analysis comprised 204 spectral channels. To reduce the memory requirement and to enhance the spatial features among grain kernels per image, the hyperspectral images were cropped into smaller images by a window of $128 \times 128$ pixels with an overlap of $64$ pixels.  The cropped images where the pixel density of grain kernels was less than half of the total pixel density of the image were removed. These cropped images were then used to build the train and test sets for our experiments.

The images were captured in both dense and sparse settings. The imaging camera  (Specim FX17 line scan camera \cite{specim}) features 640 spatial pixels and operates within the 900–1700 nm range,  detecting 224 evenly distributed spectral channels. Due to low sensitivity and noise, the first and last 10 spectral channels were excluded, leaving 204 channels for analysis. Hyperspectral images were cropped into $128 \times 128$ pixel windows with a $64$ pixel overlap to reduce memory usage and enhance spatial features. The cropped images with grain kernel pixel density less than half of the total were discarded. The remaining images formed the training and test sets for the experiments.

For training database, 360 images from each of the eight classes were randomly selected, forming a total of 2880 images. The rest of images were used to build the test database.

\subsection{Experiment settings}

For the feature embedding, we have considered popular CNN architecture ResNet-18 as backbone model. Rather than being directly inputted into the feature embedding model, the hyperspectral images undergo processing through a linear downsample layer, which is added on top of the CNN architecture as per \cite{dreier2022hyperspectral}. This modification allows for the handling of hyperspectral image data within a 2D CNN framework. The database contains a total of 8 classes.

We conducted our experiments using two scenarios as follows.

\begin{itemize}
	\item Complete class training ($8$-way classification): This approach involves training the classifier with all eight classes in the database. The objective is to evaluate the classifier’s performance on unseen images from the same classes it was trained on.
	
	\item Partial class training ($6$-way classification): In this approach, the classifier is trained using only six of the eight classes, excluding two classes from the training process. The goal is to assess how well the classifier generalises to unseen test images from the excluded classes, thus evaluating its ability to handle previously unseen classes.
\end{itemize}

We used a shot size of 5 for support sets and 10 for query sets in our experiments. The training database had 2880 images, with 360 per class, divided into 24 episodes of support and query sets. The network was trained for 50 epochs, using all 24 episodes in each. Our objective was to adapt the ResNet-18 backbone from the RGB to the HSI domain for better feature extraction from hyperspectral images. This adaptation improved similarity measurements between support and query images, enhancing classification accuracy.

Processing data with 204 channels is computationally intensive. In \cite{engstrom2023improving} and \cite{dreier2022hyperspectral}, the number of channels was reduced by 50\% and 66.66\%, respectively, by averaging every second and third consecutive channels before passing the data to the spectral downsampling layer. However, averaging hyperspectral channels can lead to information loss. Therefore, we utilised all 204 channels in our experiments.

%We conducted experiments using HSI data with 204 channels through two configurations. In the first configuration, we directly passed the 204 channel data to the spectral downsampling layer. In the second configuration, we introduced a channel attention layer (SE block) before feeding the data into the spectral downsampling layer. In addition, we conducted experiments with a third configuration using 102 channels, as suggested in \cite{engstrom2023improving}, to compare and investigate whether the channel attention mechanism improves performance or if performance degrades due to information loss from channel averaging. The experiments have been performed on two 16 GB NVIDIA GeForce RTX 4060 GPUs with a 32 GB RAM.

We conducted experiments on HSI data with 204 channels in two configurations. The first fed the 204 channels directly into the spectral downsampling layer. The second added a channel attention layer (SE block) before the spectral downsampling. In addition, we tested a third setup with 102 channels, as suggested in \cite{engstrom2023improving}. we aimed to assess whether the channel attention mechanism improves performance or if channel averaging causes information loss. The experiments were performed on a 16 GB NVIDIA GeForce RTX 4060 GPU machine with 32 GB RAM.

%Therefore, spectral reduction can increase the efficiency. We reduced the channel numbers by 50\% and 66.66\% as per \cite{engstrom2023improving} and \cite{dreier2022hyperspectral} by averaging every second and third consecutive channels, respectively. For the fine-tuning, the entire database is partitioned into training and testing sets, comprising 20\% and 80\% of the images, respectively. %Instead of directly feeding to the feature embedding model, the hyperspectral images are passed through a linear down sample layer that is added on top of the CNN architecture to accommodate the hyperspectral image processing in 2D CNN setting.

\section{Results}
%In this section, we discuss the results of our experiments. To evaluate the results on test images for the classifiers trained in 8-way, we used CCPs instead of support sets during the inference. On the other hand, to evaluate the results on images from unknown classes using the classifier trained in 6-way, we provide a support set of the unkown classes (shot size -5) during the inference.
In this section, we present the results of our experiments. The results for Complete class training and Partial class training approaches are provided in separate subsections. 

\subsection{Complete class training evaluation results}

\begin{figure}
	\centering
	\includegraphics[width=1\linewidth]{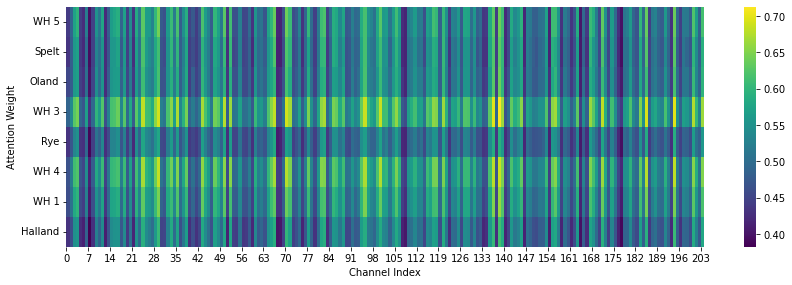}
	\caption{Heatmap for attention weight distribution}
	\label{fig:heatmap}
\end{figure}

To evaluate the classifier's performance in the 8-way classification scenario, we used CCPs instead of support sets during inference, as shown in Table \ref{8-way}. The CCPs were calculated by averaging class prototypes from each support set in the training database, as explained in Section IV. The results show improved classifier performance when all channels are used with the channel attention mechanism. This is because the hyperspectral data channels were weighted by attention based on their relevance before being processed by ResNet-18. Fig. \ref{fig:heatmap} displays a heatmap of attention weights across the 204 channels for the eight classes at the end of training.

Among the two configurations without channel attention, the reduced channel configuration performs worst, likely due to information loss from channel averaging. Table \ref{8-way-time} shows the training time for 50 epochs across the three configurations. The all-channel configuration (with attention) took the longest, while the reduced channel configuration (without attention) was the quickest. As shown in Table \ref{8-way}, the reduced channel configuration has the lowest accuracy, with differences of 2.77\% compared to the all-channel configuration (without attention) and 4.42\% compared to the all-channel configuration (with attention). Thus, for a faster classifier with minimal accuracy loss, the reduced channel configuration can be considered.

\begin{table}
	\caption{ Classification Results of the complete class training Using CCP for Inference}
	\label{8-way}
	\centering
	\begin{tabular}{ccc}
		\hline
		No of channels & Channel attention & Accuracy(\%)   \\
		\hline
		102 & N  & 93.33  \\
		204 & N & 96.10   \\ %94.16; 86.66 when inferred with 6-way classifier 
		204 & Y & 97.75   \\
		\hline
	\end{tabular}
\end{table}

\begin{table}
	\caption{ Time required for complete class training}
	\label{8-way-time}
	\centering
	\begin{tabular}{ccc}
		\hline
		No of channels & Channel attention & Time(Hours)   \\
		\hline
		102 & N  & 7.16  \\
		204 & N & 19.07   \\ %94.16; 86.66 when inferred with 6-way classifier 
		204 & Y & 22.21   \\
		\hline
	\end{tabular}
\end{table}

%In this section, we discuss the advantages of using CCPs instead of support sets with a quantitative analysis. In the Section III, we have discussed the intuitive benefits of using CCPs in terms of effectiveness as CCPs are comparatively more immune to the presence of outliers than an individual support set. Here, we present the results in Table \ref{ccp-advantage} to support that argument. The results in Table \ref{ccp-advantage} are obtained considering the experiment scenario when all channels are utilised in conjunction with the channel attention mechanism. After the classifier is being trained. We performed evaluation of the test database for each of the 24 support sets in the train database. In addition, we also evaluated the test database using the CCP. From the results, we can clearly see that, the accuracies on the test database using individual support sets are not consistent. For some support set, the accuracy is low but for some it is higher. However, the accuracy on test database using CPPs is the highest. This is because, individual support sets may have reference images that may not represent the test images from the corresponding classes effectively. However, CCPs contain the robust representation of each class and help in better classifying test images.

In this section, we present a quantitative analysis to demonstrate the advantages of using CCPs over individual support sets. In Section IV, we discussed the intuitive benefits of CCPs in terms of robustness, as CCPs are more resistant to the influence of outliers compared to individual support sets. Table \ref{ccp-advantage} provides results that substantiate this argument. These results were obtained under the experimental setup where all channels were utilised alongside a channel attention mechanism.

%After training the classifier, we evaluated the test set using each of the 24 support sets from the training database, as well as using the CCPs. The results indicate that the accuracy on the test set varies across individual support sets with a average accuracy of $96.29\% \pm 1.21\% $. Some support sets yield lower accuracies, while others achieve comparatively higher results. However, the accuracy when using CCPs is the highest (97.75\%). Therefore, by using CCP, the classifier's performance is improved 1.46\% and it is significant. This can be attributed to the fact that individual support sets may include reference images that do not effectively represent the corresponding test images. In contrast, CCPs provide a more robust representation of each class, leading to improved classification performance on the test set.

After training the classifier, we evaluated the test set using 24 support sets from the training database and the CCPs. The results show that the accuracy on the test set varies across support sets, averaging $96.29\% \pm 1.21\% $.  While some sets perform poorly, others achieve higher accuracy. Using CCPs yields the best accuracy at  97.75\%, improving performance by 1.46\%, which is significant. This improvement is likely because individual support sets may contain reference images that poorly represent test images, whereas CCPs offer a more robust representation, enhancing classification performance.

\begin{table}
	\caption{ Advanatge of using CCP over support sets}
	\label{ccp-advantage}
	\centering
	\begin{tabular}{ccc}
		\hline
		Accuracy without CCP(\%)   & Accuracy(\%) with CCP  \\
		\hline
		$96.29 \pm 1.21 $  &  97.75 \\ %0.86
		
		\hline
	\end{tabular}
\end{table}

\begin{table}
	\caption{ Comparison of classification results with \cite{dreier2022hyperspectral}}
	\label{comparison-table}
	\centering
	\begin{tabular}{cc}
		\hline
		Best accuracy (\%) from \cite{dreier2022hyperspectral} & Our best accuracy (\%)   \\
		\hline
		99.75  &  97.75 \\
		
		\hline
	\end{tabular}
\end{table}

%At inference, when paired with CCPs, the few-shot classifier trained on the complete class scenario effectively mimics a fully trained or fine-tuned supervised classifier. This is because both classifiers are tasked with predicting test images from classes they have encountered during training and the classifiers only required to extract the features of test images (in case of FSL, no need to extract features of support set images as CCPs are available).

Now, we compare the effectiveness of the FSL classifier in the complete class training scenario with a fully supervised classifier, as detailed in \cite{dreier2022hyperspectral}, which uses the same database as our experiments. Table \ref{comparison-table} presents the best classification results from \cite{dreier2022hyperspectral} alongside those from our $8$-way classification approach. During inference, when combined with CCPs, the FSL classifier trained on the complete class scenario effectively mimics a fully trained or fine-tuned supervised classifier. Both predict test images from previously seen classes, with the FSL classifier requiring no feature extraction for support set images, as CCPs are pre-available.

\begin{figure}
	\centering
	\includegraphics[width=1\linewidth]{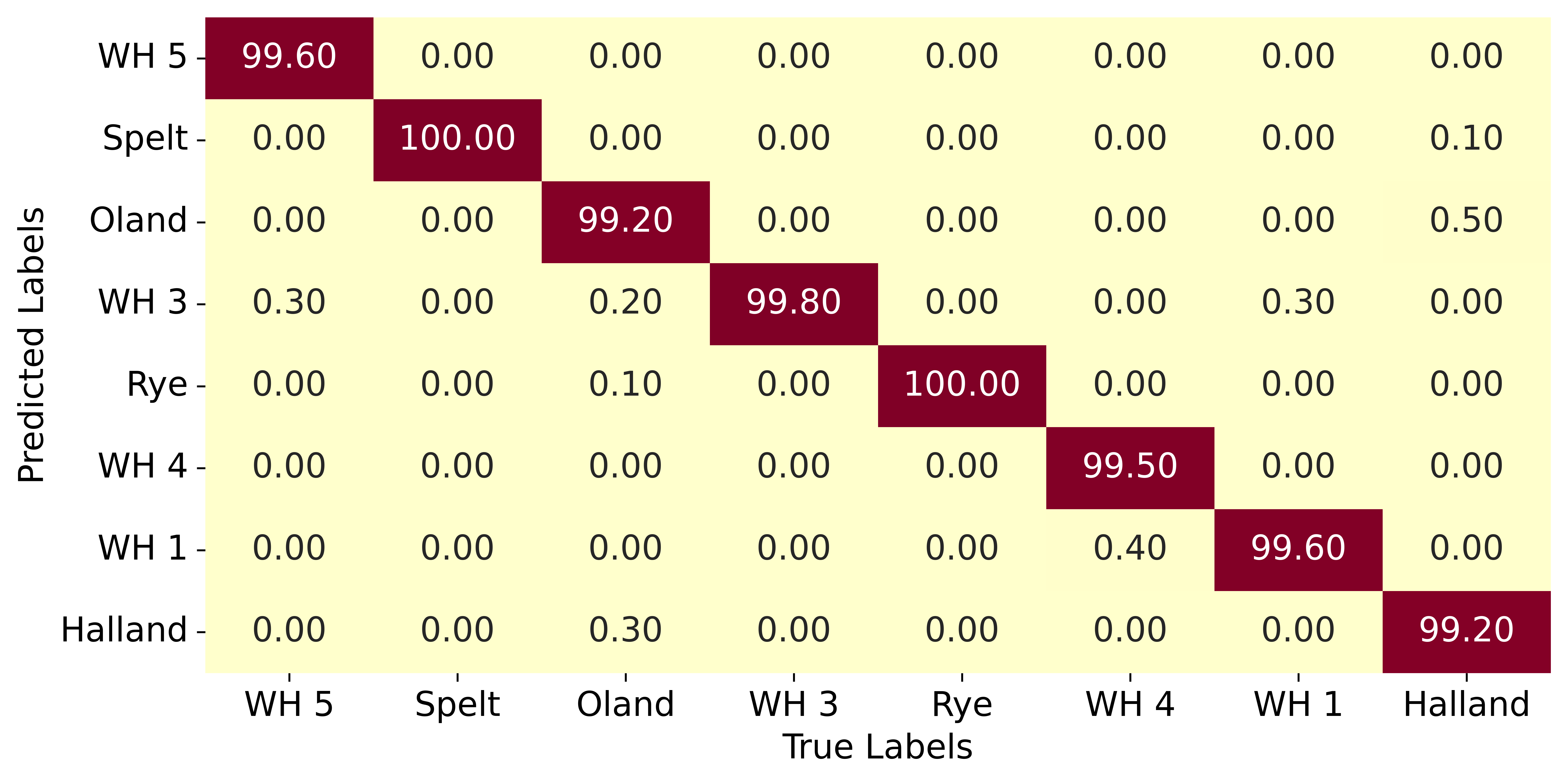}
	\caption{Confusion matrix of the best result reported in \cite{dreier2022hyperspectral}}
	\label{fig:denmark-consfusion}
\end{figure}

\begin{figure}
	\centering
	\includegraphics[width=1\linewidth]{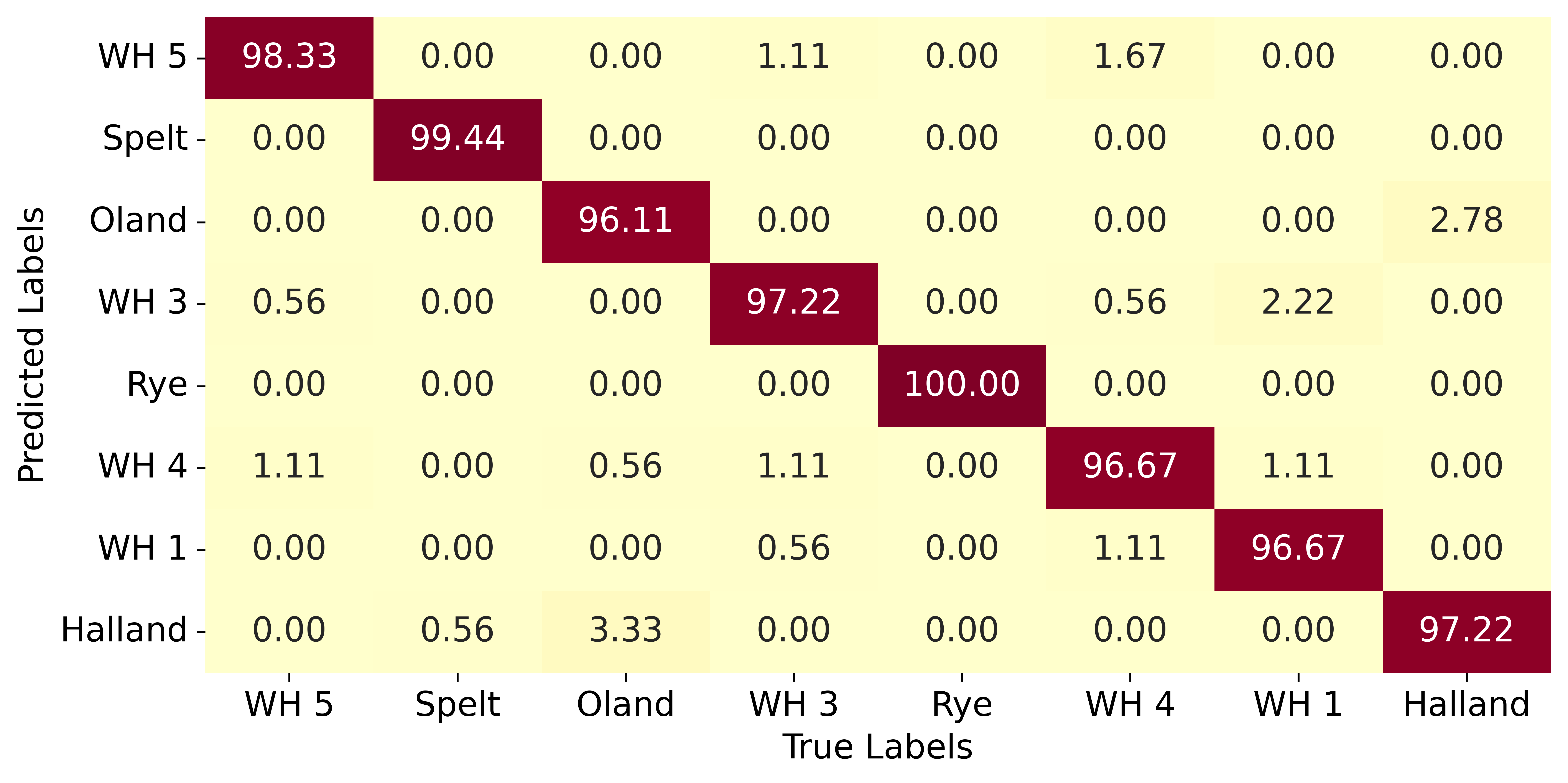}
	\caption{Confusion matrix of the best results obtained using proposed 8-way classification}
	\label{fig:our-confusion}
\end{figure}

\begin{figure}
	\centering
	\includegraphics[width=1\linewidth]{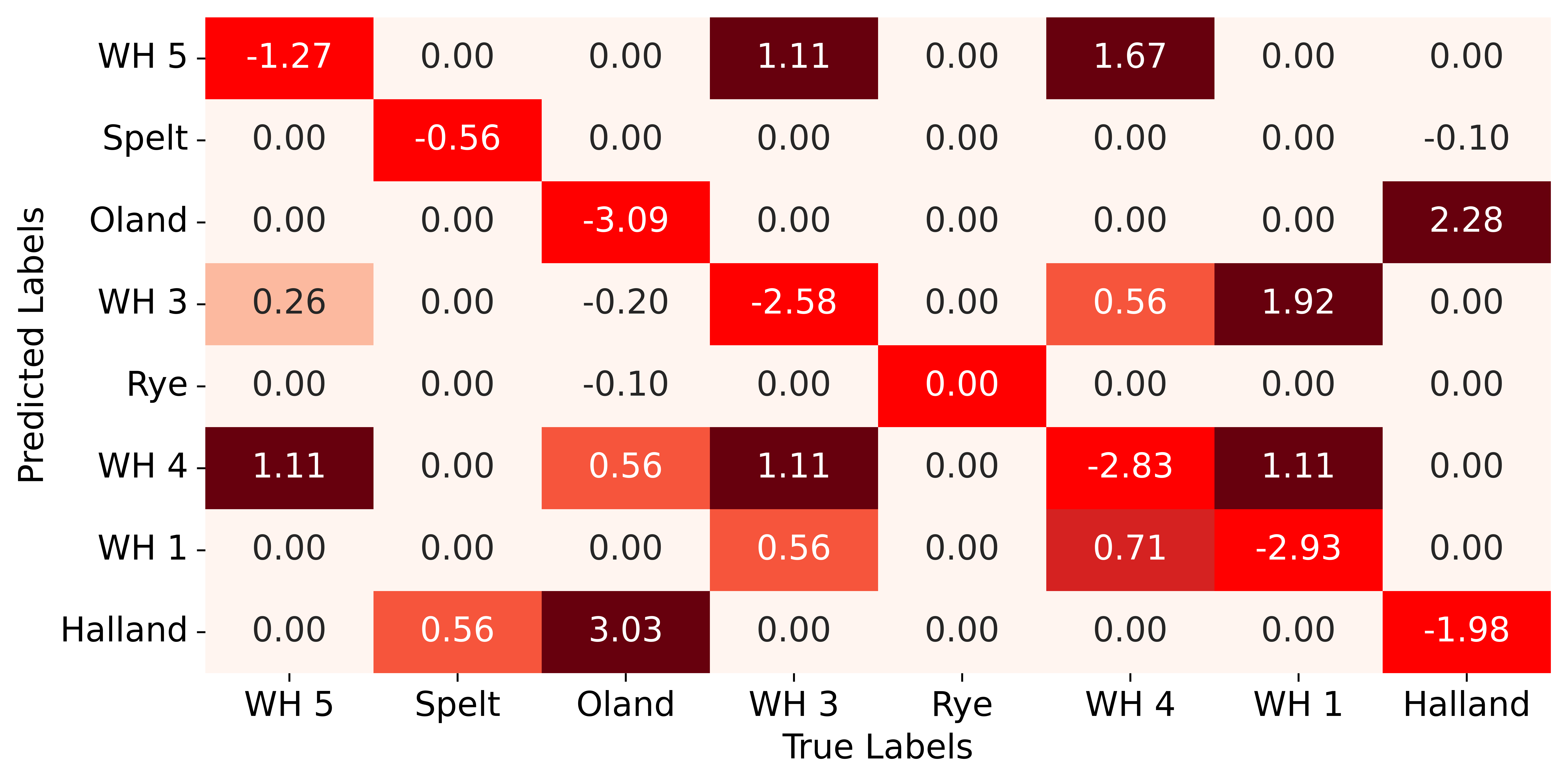}
	\caption{Confusion matrix representing the difference in accuracies}
	\label{fig:diff-confusion}
\end{figure}

A key distinction, however, lies in the size of the training databases used by the two approaches. For the complete class training of our FSL model, we utilised a training set with only 2,880 images. In contrast, the fully fine-tuned ResNet-18 model in \cite{dreier2022hyperspectral} used a much larger training set with 16,666 images. This means that our FSL model was trained using only 17.28\% of the data used in \cite{dreier2022hyperspectral}. The ability to train effectively with such a reduced database is a significant advantage in real-world scenarios, where acquiring large quantities of labelled HSI data can be challenging and resource-intensive.

Despite using a much smaller portion of the training data compared to \cite{dreier2022hyperspectral}, our FSL classifier achieves a classification accuracy of 97.75\%, which is remarkably close to the 99.75\% accuracy reported for the fully fine-tuned ResNet-18. This demonstrates the potential of FSL to achieve competitive performance with far fewer labelled samples, making it a highly practical approach for HSI tasks where data scarcity is a common issue.

%We have also provided the confusion matrix plots of the best result reported in \cite{dreier2022hyperspectral} and the best result obtained using our proposed 8-way classification approach in Figures \ref{fig:denmark-consfusion} and \ref{fig:denmark-consfusion}, respectively. We noticed that, using our proposal, the individual accuracies of three grains (i.e., Rye, Spelt and WH 5) are very close to what it is in \cite{dreier2022hyperspectral}. However, for the other five grain types, the individual accuracies are slightly low as compared to what it is in \cite{dreier2022hyperspectral} but the degradation is only in the range of (2-3)\%.  In our future work, we will bring this accuracy gap even lower making the overall accuracy more close to as it is in \cite{dreier2022hyperspectral}.

We present confusion matrix plots for the best result from \cite{dreier2022hyperspectral}, our optimal result from the proposed $8$-way classification method, and the difference in classification accuracies between our approach and that of \cite{dreier2022hyperspectral} in Figs. \ref{fig:denmark-consfusion}, \ref{fig:our-confusion} and \ref{fig:diff-confusion}, respectively. These confusion matrices show percentage values instead of actual counts. In Fig. \ref{fig:diff-confusion},  the diagonal cells represent the accuracy differences for individual grain classes to illustrate how our proposal compares to \cite{dreier2022hyperspectral} in classifying these grains.

We observed that for three grains (Rye, Spelt, and WH 5), the individual accuracies in our method are very close to those reported in \cite{dreier2022hyperspectral}. For the other five grain types, however, our method yields slightly lower accuracies, with a reduction of only about 2–3\%. In future work, we aim to reduce this accuracy gap further, bringing our overall performance closer to that of \cite{dreier2022hyperspectral}.

%Now, we discuss and analyse the performance of our proposed $8$-way classifier with the help of a 2D scatter plot as shown in Fig. \ref{fig:t-sne} that represents the 2D visualisation of prototypes of eight classes and their corresponding CCPs. Since prototypes and CCPs are high-dimensional vectors, direct visualisation in 2D space is not feasible. To address this, we applied t-distributed Stochastic Neighbor Embedding (t-SNE) \cite{van2008visualizing} for dimensionality reduction. The resulting 2D scatter plot provides a visual interpretation of the data. It is important to note that the two dimensions generated by t-SNE do not correspond to specific or interpretable features. Instead, they represent a non-linear transformation of the original high-dimensional space, where the distances between points reflect the relative similarities between the feature embeddings of the eight classes. In Fig. \ref{fig:t-sne}, small circular dots represent the prototypes from each of the 24 support sets from the training database and star symbols represent the corresponding CCPs.

\begin{figure}
	\centering
	\includegraphics[width=1\linewidth]{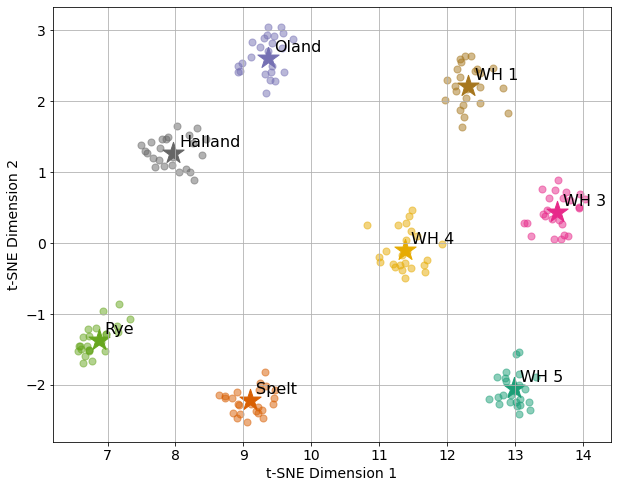}
	\caption{Visualisation of prototypes and CCPs in 2D space}
	\label{fig:t-sne}
\end{figure}

Now, we analyse the performance of our proposed $8$-way classifier using the 2D scatter plot in Fig. \ref{fig:t-sne}, which visualises prototypes of eight classes and their CCPs. As prototypes and CCPs are high-dimensional vectors, we applied t-distributed Stochastic Neighbor Embedding (t-SNE) \cite{van2008visualizing} for dimensionality reduction. This 2D plot reflects the relative similarities between feature embeddings through a non-linear transformation of the high-dimensional space. Note that the two dimensions do not correspond to specific features. In Fig. \ref{fig:t-sne}, small dots represent prototypes from 24 support sets, while star symbols denote the corresponding CCPs.

Fig. \ref{fig:t-sne} helps us understand why certain grains have higher misclassification rates in our $8$-way classifier. For instance, Oland has the lowest accuracy (96.11\%) because, in the feature embedding space, it is very close to Halland. This proximity causes 3.33\% of Oland samples to be misclassified as Halland, and similarly, 2.78\% of Halland samples to be misclassified as Oland. Likewise, WH 1 is positioned close to WH 3 and WH 4, leading to 2.22\% of WH 1 samples being misclassified as WH 3 and 1.11\% as WH 4.

\subsection{Partial class training evaluation results}
%In this section, we evaluate the performance of our few-shot classifier on two image classes excluded from training. For this purpose, we trained the classifier only for the scenario when all hyperspectral channels were used in combination with the channel attention as this is the best out of three scenarios as discussed in the section V.A using the results of Table \ref{8-way}. The two classes that are excluded from the training are Rye and WH 5. The reasons for selecting them for the exclusion list are because Rye is a different type of grain than wheat. So it would be worth checking the performance of the classifier to classify Rye when samples from Rye are not in the training set. In addition, we selected WH 5 in the exclusion list because, its protein content is least among the what varities and close to the protien content of Rye.

%In this section, we evaluate the performance of our few-shot classifier trained as per the partial class training scenario. In this scenario, two image classes, Rye and WH 5 were excluded from the training set. We performed this experiment using the best-performing configuration, where all hyperspectral channels were utilised in conjunction with the channel attention mechanism, as identified in Section VI.A based on the results in Table \ref{8-way}.

In this section, we assess the performance of our few-shot classifier under the partial class training scenario, excluding Rye and WH 5 from the training set. Using the optimal configuration identified in Section VI.A (Table \ref{8-way}), all hyperspectral channels were utilised in conjunction with the channel attention mechanism for this experiment.

%The selection of Rye and WH 5 for exclusion was intentional. Rye was excluded because it is a different type of grain compared to wheat, making it an ideal candidate to assess how well the classifier can generalise to unseen classes that differ significantly from the training set. Evaluating the classifier’s ability to correctly classify Rye when no Rye samples were used during training allows us to measure its robustness in handling domain shifts. We also excluded WH 5 because it has the lowest protein content among the wheat varieties, and its protein content is closer to that of Rye. 

Rye and WH 5 were excluded intentionally. Rye, being a different grain type, was ideal for testing the classifier’s ability to generalise to unseen classes. Evaluating its classification without Rye samples in training assesses robustness to domain shifts. WH 5, with the lowest protein content among wheat varieties, was excluded as its protein content closely resembles that of Rye.

To assess the classifier's performance on test images from excluded classes, we used individual support sets instead of CCPs. This simulates real-world scenarios where classifiers must handle unseen classes not included in training. As the excluded classes were absent from the training database, generating CCPs for them post-training was impossible. In practical settings, labelled data for unseen classes is often scarce, with only a few support sets available for inference. Since CCPs require multiple support sets to form robust prototypes, our method of using individual support sets mirrors realistic conditions, emphasising the classifier's ability to generalise with minimal labelled data.

By analysing the performance on these excluded classes, we aim to gain insight into the generalisation ability and class discrimination power of the classifier, especially in handling out-of-distribution and closely related classes.

%The goal is to assess how well the classifier generalises to the unseen classes during inference.%, simulating a real-world scenario where new, previously unseen categories must be recognised. 
We employed two different evaluation strategies to investigate the performance of the classifier on the excluded classes. The results are provided in Table \ref{6-way}.

\begin{table}
	\caption{ Performance of the 6-Way Classifier on Unseen Classes with Support Sets for Inference}
	\label{6-way}
	\raggedright
	\begin{tabular}{cc}
		\hline
		Strategy &  Accuracy(\%)   \\
		\hline
		\multicolumn{1}{l}{Strategy 1: Support Set with Only Excluded Classes} & 98.33    \\
		\multicolumn{1}{l}{Strategy 2: Support Set with All Classes} & 83.89    \\
		%204 & Y &   98.33\\  %83.89 %79.98 when support set has other classes. total 8 classes. 
		\hline
	\end{tabular}
\end{table}

\subsubsection{Strategy 1: Support Set with Only Excluded Classes}

In this strategy, the support set contains only the images from the two excluded classes. 
%For the classifier trained with the 6-way approach, we assessed its performance on images from the unseen classes by providing a support set of the unknown classes (with a shot size of 5) during inference.
The classifier is therefore tasked with distinguishing between only these two unseen classes during inference. This setting isolates the model's ability to discriminate between new classes, without interference from the training set classes.

The result for this strategy show a high accuracy of 98.33\%, indicating that the classifier can effectively recognise and differentiate between the excluded classes when given a limited number of comparison categories.  This high accuracy highlights the effectiveness of learned few-shot representations for unseen classes in controlled settings. With only two categories, the reduced complexity increases success rates, making this approach ideal for scenarios with a limited number of known classes, such as binary decision systems or classification tasks within specific subcategories.

%This high accuracy demonstrates the effectiveness of the learned few-shot representations for unseen classes in a controlled environment. Since the classifier is only choosing between two possible categories, the problem complexity is greatly reduced, resulting in a higher success rate. This setting is useful in scenarios where the number of possible classes is known and limited during inference. For example, applications in specific domain tasks where only a narrow set of possible outcomes is expected (e.g., binary decision systems or classification tasks within specific subcategories).

%Limitation: The key limitation of this approach is that it does not reflect real-world conditions, where the classifier would need to differentiate between a much larger set of possible classes, both seen and unseen.

\subsubsection{Strategy 2: Support Set with All Classes (Excluded + Trained Classes)}

In the second strategy, the support set includes images from all eight classes, i.e., it includes both the six classes the classifier was trained on and the two excluded classes. The goal here is to simulate a real-world scenario where the classifier encounters a larger set of potential candidates during inference, as it must compare both the excluded and trained classes.

In this setting, the classifier achieved an accuracy of 83.89\%. The decrease in performance compared to Strategy 1 (98.33\%) is expected, as the complexity of the task increases when the classifier needed to distinguish between a broader set of eight classes, not just two.

This result provides a more realistic evaluation of the classifier’s generalisation capabilities. In real-world applications, classifiers will need to deal with an open-world environment where both known (trained) and unknown (unseen) categories may present during inference. The 83.89\% accuracy in this scenario suggests that while the classifier can still generalise to unseen classes, the presence of a larger number of comparison classes introduces additional complexity that naturally reduces its ability to accurately classify the excluded classes.

%Advantage: This approach better simulates practical applications where the system encounters a diverse set of classes, including those from training and potentially new classes in the wild. It demonstrates how the classifier generalizes beyond its training set while facing increased decision complexity.

%\subsubsection{Comparative Analysis and Real-World Relevance}
\subsubsection{Comparative Analysis}

\begin{figure}
	\centering
	\includegraphics[width=1\linewidth]{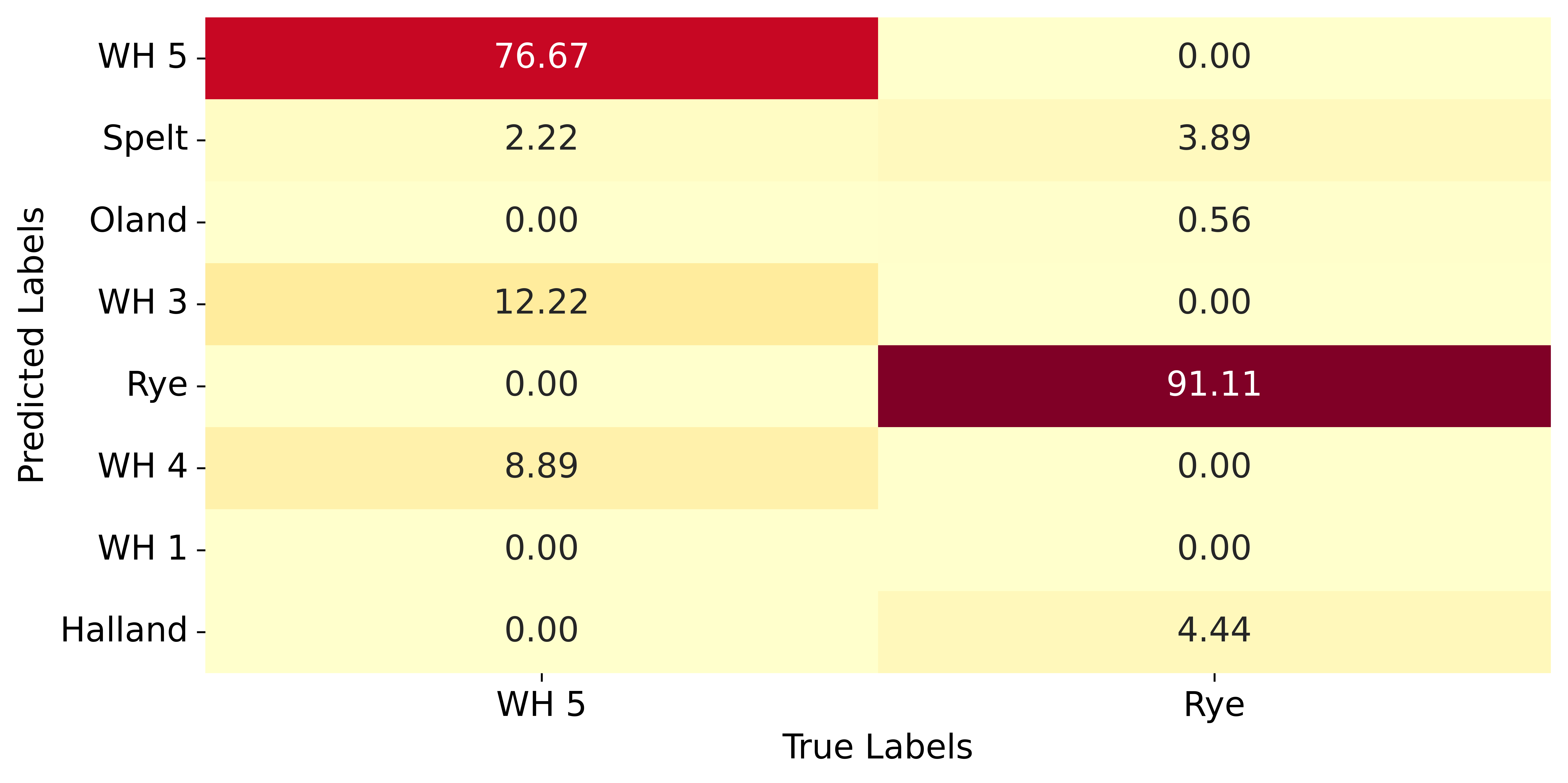}
	\caption{Confusion matrix plot of the evaluation results from partial class training}
	\label{fig:cm}
\end{figure}

In this section, we investigate the probable reason of the classifier's under performance in Strategy 2. We have presented a confusion matrix plot in Fig \ref{fig:cm} from the results obtained by Strategy 2. 
As shown in Fig. \ref{fig:cm}, the classification accuracy of Rye is significantly higher than that of WH 5 among the two excluded classes. This is because, in the feature embedding space (refer to Fig. \ref{fig:t-sne}), Rye is more distant from most classes, whereas WH 5 is positioned closer to WH 4 and WH 3. Consequently, in the presence of all classes, WH 5 exhibits a higher misclassification frequency compared to Rye, with the majority of misclassifications occurring for classes in close proximity to WH 5 in the feature embedding space, specifically WH3 (12.22\% misclassification rate) and WH 4 (8.89\% misclassification rate).

\section{Conclusion}

This paper explored FSL scenarios for hyperspectral image classification in grain quality assessment, yielding significant findings and contributions, specifically when collecting labelled data is difficult. We introduced CCPs as a robust alternative to individual support sets during inference, demonstrating superior performance and reduced susceptibility to outliers. Our proposed channel attention mechanism enhanced feature representation of hyperspectral images, resulting in improved classification performance. Notably, the FSL approach achieved comparable accuracy to fully trained supervised classifiers while using significantly less training data, highlighting its potential in scenarios where labelled data is scarce.

Our research has important implications for hyperspectral image classification in supply chain applications, specifically in terms of reduced data requirements and adaptability to unseen classes. The classifier's ability to generalise to unseen classes is crucial for applications where new grain grades or types may be encountered. Future research could focus on exploring more complex FSL architectures, investigating applications in other domains within hyperspectral image classification, and developing strategies to enhance performance when dealing with closely related classes in the feature embedding space. Overall, this paper demonstrates the potential of FSL in hyperspectral image classification for grain quality assessment, offering a promising approach that balances accuracy, efficiency, and adaptability in real-world supply chain applications.

\bibliographystyle{IEEEtai}
\bibliography{hsi_classification_ref.bib}	

\begin{thebibliography}{10}
\providecommand{\url}[1]{#1}
\csname url@samestyle\endcsname
\providecommand{\newblock}{\relax}
\providecommand{\bibinfo}[2]{#2}
\providecommand{\BIBentrySTDinterwordspacing}{\spaceskip=0pt\relax}
\providecommand{\BIBentryALTinterwordstretchfactor}{4}
\providecommand{\BIBentryALTinterwordspacing}{\spaceskip=\fontdimen2\font plus
\BIBentryALTinterwordstretchfactor\fontdimen3\font minus
  \fontdimen4\font\relax}
\providecommand{\BIBforeignlanguage}[2]{{%
\expandafter\ifx\csname l@#1\endcsname\relax
\typeout{** WARNING: IEEEtran.bst: No hyphenation pattern has been}%
\typeout{** loaded for the language `#1'. Using the pattern for}%
\typeout{** the default language instead.}%
\else
\language=\csname l@#1\endcsname
\fi
#2}}
\providecommand{\BIBdecl}{\relax}
\BIBdecl

\bibitem{ariana2006near}
D.~P. Ariana, R.~Lu, and D.~E. Guyer, ``Near-infrared hyperspectral reflectance
  imaging for detection of bruises on pickling cucumbers,'' \emph{Computers and
  electronics in agriculture}, vol.~53, no.~1, pp. 60--70, 2006.

\bibitem{zhu2019rapid}
X.~Zhu and G.~Li, ``Rapid detection and visualization of slight bruise on
  apples using hyperspectral imaging,'' \emph{International journal of food
  properties}, vol.~22, no.~1, pp. 1709--1719, 2019.

\bibitem{he2015hyperspectral}
H.-J. He and D.-W. Sun, ``Hyperspectral imaging technology for rapid detection
  of various microbial contaminants in agricultural and food products,''
  \emph{Trends in Food Science \& Technology}, vol.~46, no.~1, pp. 99--109,
  2015.

\bibitem{calvini2020exploring}
R.~Calvini, S.~Michelini, V.~Pizzamiglio, G.~Foca, and A.~Ulrici, ``Exploring
  the potential of nir hyperspectral imaging for automated quantification of
  rind amount in grated parmigiano reggiano cheese,'' \emph{Food Control}, vol.
  112, p. 107111, 2020.

\bibitem{feng2020colour}
C.-H. Feng and Y.~Makino, ``Colour analysis in sausages stuffed in modified
  casings with different storage days using hyperspectral imaging--a
  feasibility study,'' \emph{Food Control}, vol. 111, p. 107047, 2020.

\bibitem{shuqin2016predicting}
Y.~Shuqin, H.~Dongjian, and N.~Jifeng, ``Predicting wheat kernels’ protein
  content by near infrared hyperspectral imaging,'' \emph{International Journal
  of Agricultural and Biological Engineering}, vol.~9, no.~2, pp. 163--170,
  2016.

\bibitem{mahesh2011identification}
S.~Mahesh, D.~Jayas, J.~Paliwal, and N.~White, ``Identification of wheat
  classes at different moisture levels using near-infrared hyperspectral images
  of bulk samples,'' \emph{Sensing and instrumentation for food quality and
  safety}, vol.~5, pp. 1--9, 2011.

\bibitem{barbedo2018detection}
J.~G. Barbedo, E.~M. Guarienti, and C.~S. Tibola, ``Detection of sprout damage
  in wheat kernels using nir hyperspectral imaging,'' \emph{Biosystems
  Engineering}, vol. 175, pp. 124--132, 2018.

\bibitem{liang2020comparison}
K.~Liang, J.~Huang, R.~He, Q.~Wang, Y.~Chai, and M.~Shen, ``Comparison of
  vis-nir and swir hyperspectral imaging for the non-destructive detection of
  don levels in fusarium head blight wheat kernels and wheat flour,''
  \emph{Infrared Physics \& Technology}, vol. 106, p. 103281, 2020.

\bibitem{kniese2021classification}
J.~Kniese, A.~M. Race, and H.~Schmidt, ``Classification of cereal flour species
  using raman spectroscopy in combination with spectra quality control and
  multivariate statistical analysis,'' \emph{Journal of Cereal Science}, vol.
  101, p. 103299, 2021.

\bibitem{neethirajan2007detection}
S.~Neethirajan, D.~S. Jayas, and N.~White, ``Detection of sprouted wheat
  kernels using soft x-ray image analysis,'' \emph{Journal of Food
  Engineering}, vol.~81, no.~3, pp. 509--513, 2007.

\bibitem{bao2019rapid}
Y.~Bao, C.~Mi, N.~Wu, F.~Liu, and Y.~He, ``Rapid classification of wheat grain
  varieties using hyperspectral imaging and chemometrics,'' \emph{Applied
  Sciences}, vol.~9, no.~19, p. 4119, 2019.

\bibitem{majumdar2000classification}
S.~Majumdar and D.~S. Jayas, ``Classification of cereal grains using machine
  vision: Iv. combined morphology, color, and texture models,''
  \emph{Transactions of the ASAE}, vol.~43, no.~6, pp. 1689--1694, 2000.

\bibitem{pourreza2012identification}
A.~Pourreza, H.~Pourreza, M.-H. Abbaspour-Fard, and H.~Sadrnia,
  ``Identification of nine iranian wheat seed varieties by textural analysis
  with image processing,'' \emph{Computers and electronics in agriculture},
  vol.~83, pp. 102--108, 2012.

\bibitem{gowen2007hyperspectral}
A.~A. Gowen, C.~P. O'Donnell, P.~J. Cullen, G.~Downey, and J.~M. Frias,
  ``Hyperspectral imaging--an emerging process analytical tool for food quality
  and safety control,'' \emph{Trends in food science \& technology}, vol.~18,
  no.~12, pp. 590--598, 2007.

\bibitem{pub.1064892569}
\BIBentryALTinterwordspacing
T.~Bramble, F.~E. Dowell, and T.~J. Herrman, ``Single-kernel near-infrared
  protein prediction and the role of kernel weight in hard red winter wheat,''
  \emph{Applied Engineering in Agriculture}, vol.~22, no.~6, pp. 945--949,
  2006. [Online]. Available:
  \url{https://app.dimensions.ai/details/publication/pub.1064892569}
\BIBentrySTDinterwordspacing

\bibitem{qiu2018variety}
Z.~Qiu, J.~Chen, Y.~Zhao, S.~Zhu, Y.~He, and C.~Zhang, ``Variety identification
  of single rice seed using hyperspectral imaging combined with convolutional
  neural network,'' \emph{Applied Sciences}, vol.~8, no.~2, p. 212, 2018.

\bibitem{weng2020hyperspectral}
S.~Weng, P.~Tang, H.~Yuan, B.~Guo, S.~Yu, L.~Huang, and C.~Xu, ``Hyperspectral
  imaging for accurate determination of rice variety using a deep learning
  network with multi-feature fusion,'' \emph{Spectrochimica Acta Part A:
  Molecular and Biomolecular Spectroscopy}, vol. 234, p. 118237, 2020.

\bibitem{karmakar2023guide}
P.~Karmakar, S.~W. Teng, M.~Murshed, P.~Pang, and C.~Van~Bui, ``A guide to
  employ hyperspectral imaging for assessing wheat quality at different stages
  of supply chain in australia: A review,'' \emph{IEEE Transactions on AgriFood
  Electronics}, vol.~1, no.~1, pp. 29--40, 2023.

\bibitem{ravikanth2016detection}
L.~Ravikanth, V.~Chelladurai, D.~S. Jayas, and N.~D. White, ``Detection of
  broken kernels content in bulk wheat samples using near-infrared
  hyperspectral imaging,'' \emph{Agricultural Research}, vol.~5, pp. 285--292,
  2016.

\bibitem{shao2020determination}
Y.~Shao, C.~Gao, G.~Xuan, X.~Gao, Y.~Chen, and Z.~Hu, ``Determination of
  damaged wheat kernels with hyperspectral imaging analysis,''
  \emph{International Journal of Agricultural and Biological Engineering},
  vol.~13, no.~5, pp. 194--198, 2020.

\bibitem{dreier2022hyperspectral}
E.~S. Dreier, K.~M. Sorensen, T.~Lund-Hansen, B.~M. Jespersen, and K.~S.
  Pedersen, ``Hyperspectral imaging for classification of bulk grain samples
  with deep convolutional neural networks,'' \emph{Journal of Near Infrared
  Spectroscopy}, vol.~30, no.~3, pp. 107--121, 2022.

\bibitem{zhao2019laboratory}
M.~Zhao, J.~Chen, and Z.~He, ``A laboratory-created dataset with ground truth
  for hyperspectral unmixing evaluation,'' \emph{IEEE Journal of Selected
  Topics in Applied Earth Observations and Remote Sensing}, vol.~12, no.~7, pp.
  2170--2183, 2019.

\bibitem{nalepa2021recent}
J.~Nalepa, ``Recent advances in multi-and hyperspectral image analysis,''
  \emph{Sensors}, vol.~21, no.~18, p. 6002, 2021.

\bibitem{ren2024few}
J.~Ren, C.~Li, Y.~An, W.~Zhang, and C.~Sun, ``Few-shot fine-grained image
  classification: A comprehensive review,'' \emph{AI}, vol.~5, no.~1, pp.
  405--425, 2024.

\bibitem{bai2022few}
J.~Bai, S.~Huang, Z.~Xiao, X.~Li, Y.~Zhu, A.~C. Regan, and L.~Jiao, ``Few-shot
  hyperspectral image classification based on adaptive subspaces and feature
  transformation,'' \emph{IEEE Transactions on Geoscience and Remote Sensing},
  vol.~60, pp. 1--17, 2022.

\bibitem{dvornik2019diversity}
N.~Dvornik, C.~Schmid, and J.~Mairal, ``Diversity with cooperation: Ensemble
  methods for few-shot classification,'' in \emph{Proceedings of the IEEE/CVF
  international conference on computer vision}, 2019, pp. 3723--3731.

\bibitem{li2023deep}
X.~Li, X.~Yang, Z.~Ma, and J.-H. Xue, ``Deep metric learning for few-shot image
  classification: A review of recent developments,'' \emph{Pattern
  Recognition}, vol. 138, p. 109381, 2023.

\bibitem{lee2024unlocking}
G.~Y. Lee, T.~Dam, M.~M. Ferdaus, D.~P. Poenar, and V.~N. Duong, ``Unlocking
  the capabilities of explainable few-shot learning in remote sensing,''
  \emph{Artificial Intelligence Review}, vol.~57, no.~7, p. 169, 2024.

\bibitem{finn2017model}
C.~Finn, P.~Abbeel, and S.~Levine, ``Model-agnostic meta-learning for fast
  adaptation of deep networks,'' in \emph{International conference on machine
  learning}.\hskip 1em plus 0.5em minus 0.4em\relax PMLR, 2017, pp. 1126--1135.

\bibitem{new_grade_arrival}
\BIBentryALTinterwordspacing
L.~Mathews. (2020) Gains in grains – is australia producing the most
  profitable quality of wheat? Accessed: 23/10/2024. [Online]. Available:
  \url{https://grdc.com.au/resources-and-publications/grdc-update-papers/tab-content/grdc-update-papers/2020/02/gains-in-grains-is-australia-producing-the-most-profitable-quality-of-wheat}
\BIBentrySTDinterwordspacing

\bibitem{abdul2022genetically}
M.~Abdul~Aziz, F.~Brini, H.~Rouached, and K.~Masmoudi, ``Genetically engineered
  crops for sustainably enhanced food production systems,'' \emph{Frontiers in
  plant science}, vol.~13, p. 1027828, 2022.

\bibitem{luo2023closer}
X.~Luo, H.~Wu, J.~Zhang, L.~Gao, J.~Xu, and J.~Song, ``A closer look at
  few-shot classification again,'' in \emph{International Conference on Machine
  Learning}.\hskip 1em plus 0.5em minus 0.4em\relax PMLR, 2023, pp.
  23\,103--23\,123.

\bibitem{pal2022few}
D.~Pal, V.~Bundele, R.~Sharma, B.~Banerjee, and Y.~Jeppu, ``Few-shot open-set
  recognition of hyperspectral images with outlier calibration network,'' in
  \emph{Proceedings of the IEEE/CVF Winter Conference on Applications of
  Computer Vision}, 2022, pp. 3801--3810.

\bibitem{zheng2023fusion}
L.~Zheng, M.~Zhao, J.~Zhu, L.~Huang, J.~Zhao, D.~Liang, and D.~Zhang, ``Fusion
  of hyperspectral imaging (hsi) and rgb for identification of soybean kernel
  damages using shufflenet with convolutional optimization and cross stage
  partial architecture,'' \emph{Frontiers in Plant Science}, vol.~13, p.
  1098864, 2023.

\bibitem{wang2024hypersigma}
D.~Wang, M.~Hu, Y.~Jin, Y.~Miao, J.~Yang, Y.~Xu, X.~Qin, J.~Ma, L.~Sun, C.~Li
  \emph{et~al.}, ``Hypersigma: Hyperspectral intelligence comprehension
  foundation model,'' \emph{arXiv preprint arXiv:2406.11519}, 2024.

\bibitem{xu2021motion}
F.~Xu, L.~Yu, B.~Wang, W.~Yang, G.-S. Xia, X.~Jia, Z.~Qiao, and J.~Liu,
  ``Motion deblurring with real events,'' in \emph{Proceedings of the IEEE/CVF
  International Conference on Computer Vision}, 2021, pp. 2583--2592.

\bibitem{wang2020generalizing}
Y.~Wang, Q.~Yao, J.~T. Kwok, and L.~M. Ni, ``Generalizing from a few examples:
  A survey on few-shot learning,'' \emph{ACM computing surveys (csur)},
  vol.~53, no.~3, pp. 1--34, 2020.

\bibitem{parnami2022learning}
A.~Parnami and M.~Lee, ``Learning from few examples: A summary of approaches to
  few-shot learning,'' \emph{arXiv preprint arXiv:2203.04291}, 2022.

\bibitem{kulismetric}
B.~Kulis, ``Metric learning: A survey. foundations and trends{\textregistered}
  in machine learning 5 (4), 287--364 (2013).''

\bibitem{snell2017prototypical}
J.~Snell, K.~Swersky, and R.~Zemel, ``Prototypical networks for few-shot
  learning,'' \emph{Advances in neural information processing systems},
  vol.~30, 2017.

\bibitem{hu2018squeeze}
J.~Hu, L.~Shen, and G.~Sun, ``Squeeze-and-excitation networks,'' in
  \emph{Proceedings of the IEEE conference on computer vision and pattern
  recognition}, 2018, pp. 7132--7141.

\bibitem{naresh2024empirical}
V.~Naresh, G.~Yogeswararao, R.~Malmathanraj, P.~Palanisamy \emph{et~al.},
  ``Empirical analysis of squeeze and excitation based densely connected cnn
  for chilli leaf disease identification,'' \emph{IEEE Transactions on
  Artificial Intelligence}, 2024.

\bibitem{nguyen2024hyperspectral}
X.~T. Nguyen and G.~S. Tran, ``Hyperspectral image classification using an
  encoder-decoder model with depthwise separable convolution, squeeze and
  excitation blocks,'' \emph{Earth Science Informatics}, vol.~17, no.~1, pp.
  527--538, 2024.

\bibitem{wang2023advances}
X.~Wang, J.~Liu, W.~Chi, W.~Wang, and Y.~Ni, ``Advances in hyperspectral image
  classification methods with small samples: A review,'' \emph{Remote Sensing},
  vol.~15, no.~15, p. 3795, 2023.

\bibitem{zhong2017spectral}
Z.~Zhong, J.~Li, Z.~Luo, and M.~Chapman, ``Spectral--spatial residual network
  for hyperspectral image classification: A 3-d deep learning framework,''
  \emph{IEEE Transactions on Geoscience and Remote Sensing}, vol.~56, no.~2,
  pp. 847--858, 2017.

\bibitem{zhao2024improved}
L.~Zhao and Z.~Zhang, ``A improved pooling method for convolutional neural
  networks,'' \emph{Scientific Reports}, vol.~14, no.~1, p. 1589, 2024.

\bibitem{specim}
``Specim. specim fx17,,'' \url{https://www.specim.fi/products/specim-fx17}.

\bibitem{engstrom2023improving}
O.-C.~G. Engstr{\o}m, E.~S. Dreier, B.~M. Jespersen, and K.~S. Pedersen,
  ``Improving deep learning on hyperspectral images of grain by incorporating
  domain knowledge from chemometrics,'' in \emph{Proceedings of the IEEE/CVF
  International Conference on Computer Vision}, 2023, pp. 485--494.

\bibitem{van2008visualizing}
L.~Van~der Maaten and G.~Hinton, ``Visualizing data using t-sne.''
  \emph{Journal of machine learning research}, vol.~9, no.~11, 2008.

\end{thebibliography}

\vspace{12pt}

\end{document}